\documentclass{article}
\usepackage{spconf,amsmath,graphicx}
\pdfoutput=1
\usepackage[utf8]{inputenc}
\usepackage{graphicx,subfigure}
\usepackage{mathtools}
\usepackage{amsfonts}
\usepackage[colorlinks,linkcolor=black,anchorcolor=black,citecolor=black]{hyperref}
\usepackage{caption2}
\usepackage{indentfirst}
\usepackage{cases}
\usepackage{float}
\usepackage{multirow}
\usepackage{booktabs}
\usepackage{graphicx}
\usepackage{epstopdf}

\usepackage{algorithm}
\usepackage{algorithmic}

\usepackage{multicol}
\usepackage{fancyhdr}
\usepackage{multirow}
\usepackage{subfigure}
\usepackage{setspace}
\usepackage[T1]{fontenc}
\usepackage[utf8]{inputenc}

\usepackage{floatflt}

\usepackage{url}
\usepackage{graphicx}  
\usepackage{float}  
\usepackage[misc]{ifsym}

\usepackage{xcolor}

\title{eCIL-MU: Embedding based Class Incremental Learning and Machine Unlearning}
\name{ Zhiwei Zuo\textsuperscript{1,3} \thanks{The work is supported by the National Key Research and Development Program of China (2021ZD40303), the National Natural Science Foundation of China (Grant Nos. 62225205, 92055213), Natural Science Foundation of Hunan Province of China (2021JJ10023), the Science and Technology Program of Changsha (kh2301011), Shenzhen Basic Research Project (Natural Science Foundation) (JCYJ20210324140002006). Zhiwei Zuo did this work while she was a visiting student at NTU Singapore, funded by CSC scholarship.\\This work has been submitted to the IEEE for possible publication. Copyright may be transferred without notice, after which this version may no longer be accessible.},
  Zhuo Tang\textsuperscript{1,2*}, 
  Bin Wang\textsuperscript{1},
  Kenli Li\textsuperscript{1},
  Anwitaman Datta\textsuperscript{3}
}
\address{
  \textsuperscript{1} College of Computer Science and Electronic Engineering, Hunan University, China \\
  \textsuperscript{2} Shenzhen Research Institute, Hunan University, China \\
  \textsuperscript{3} School of Computer Science and Engineering, Nanyang Technological University, Singapore
}
\begin{document}
\maketitle
\begin{abstract}

New categories may be introduced over time, or existing categories may need to be reclassified. Class incremental learning (CIL) is employed for the gradual acquisition of knowledge about new categories while preserving information about previously learned ones in such dynamic environments. It might also be necessary to also eliminate the influence of related categories on the model to adapt to reclassification. We thus introduce class-level machine unlearning (MU) within CIL. Typically, MU methods tend to be time-consuming and can potentially harm the model's performance. A continuous stream of unlearning requests could lead to catastrophic forgetting. To address these issues, we propose a non-destructive eCIL-MU framework based on embedding techniques to map data into vectors and then be stored in vector databases. Our approach exploits the overlap between CIL and MU tasks for acceleration. Experiments demonstrate the capability of achieving unlearning effectiveness and orders of magnitude (upto $\sim 278\times$) of acceleration. 

\end{abstract}

\vspace{-0.5em}
\section{Introduction}
\label{sec:intro}

Machine learning models need to rapidly adapt to new information. Class Incremental Learning (CIL) \cite{lopez2017gradient,rebuffi2017icarl, wang2021few} is crucial in this context. CIL enables models to learn multiple tasks, sequentially adding new classes while retaining prior knowledge. However, as data evolves, historical information might lose relevance due to privacy, regulations, or changing insights. Take medical diagnosis with AI as an example. Imagine disease subcategories belonging to larger categories: $x_1, x_2 \in A_1$, $y_1, y_2 \in A_2$, and a diagnostic model $M$ capable of diagnosing them. Now, with $z_1, z_2 \in B$ and $A_1\cap B = A_2 \cap B = \emptyset$, $M$ needs to learn $B$. Also, research might reclassify $y_1 \in A_1$ and $y_2 \in B$, rendering category $A_2$ obsolete. To remain accurate, models must adopt CIL and unlearn outdated categories. Machine Unlearning 
 (MU) \cite{warnecke2021machine,bourtoule2021machine,li2017learning} can help models eliminate the influence of outdated data.

To achieve CIL while unlearning classes, we integrate CIL and MU. At this setting, a model is trained on multiple tasks, learning new knowledge and unlearning certain information. Post-CIL, the model must preserve discrimination for all seen classes. Then, in MU tasks, the model concentrates on unlearning specific classes. Notably, there is non-overlapping of classes across tasks. This framework enables learning new classes while removing outdated ones.

Existing machine unlearning methods fall into two categories: model-agnostic and model-intrinsic approaches. Model-agnostic methods consider distributed unlearning, federated unlearning  \cite{fraboni2022sequential}, and verifiable machine unlearning \cite{gao2022verifi}. Model-intrinsic approaches target specific models, such as Graph Neural Networks \cite{chen2022graph}, regression \cite{wu2020priu}, and Bayesian models \cite{chen2022near}. Existing methods alter models to unlearning. However, such operations give rise to several issues: foremost, modifying a model takes an amount of time and potentially hurts the model's performance, furthermore, frequent changes to a model due to unlearning requests risk catastrophic forgetting \cite{dupuy2023quantifying}, and finally, implementing unlearning and incremental learning simultaneously is challenging.

We propose an embedding based Class Incremental Learning and Machine Unlearning (eCIL-MU) framework to address these issues. We modify data rather than the CIL-MU model leading to a non-destructive unlearning. Employing an embedding technique, we map training data into vectors and utilize vector databases \cite{guo2022manu} for vectors storage and unlearning. Specifically, during a CIL task, we store vectors in the DB-CIL, identify vectors linked to classes to be unlearned in DB-CIL and then transfer them to the unlearning database DB-MU during MU process. Vectors migration can be performed asynchronously, and the temporal overlap between CIL and MU allows model training acceleration.

With vector databases holding latent vectors, a vector filter decides whether input data fits DB-MU during inference phase. For remaining classes, the model makes
predictions as usual but unlearning class data gets partly accurate outputs.

Four output strategies are considered for unlearning class data: choose with probabilities uniformly at random, proportional to the relative frequency of classes or inversely proportional to distances, alternatively, deterministically shift to nearest class. 
\begin{figure}
    \centering
    \includegraphics[width=0.49\textwidth]{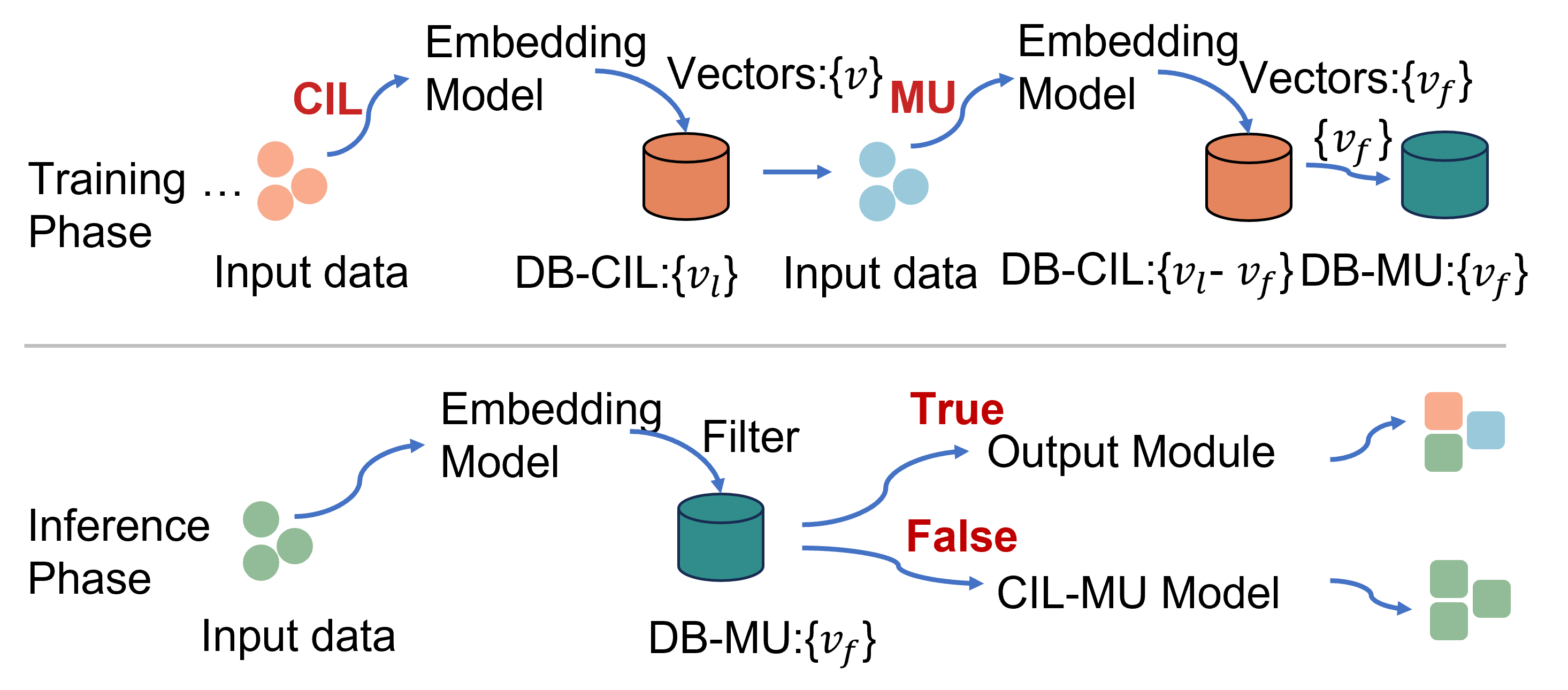}
    \vspace{-1em}
    \caption{Training phase and inference phase of eCIL-MU}
    \label{eCIL-MU}
    \vspace{-1.5em}
\end{figure}

Our main contributions are: 1. We propose the eCIL-MU framework by integrating CIL and class-level MU, in doing so, we modify data rather than the model. 2. We devise vector databases within the CIL-MU model for both learned and unlearned classes. The transfers occurs from DB-CIL to DB-MU based on cosine similarity ensuring effective migration. It exploits an overlap between CIL and MU thereby achieving acceleration. 3. We apply four distinct output strategies for the unlearning during the inference phase, and find that using the strategy shift-to-nearest-class achieves non-destructive effects, while also delaying catastrophic forgetting.

\vspace{-0.5em}
\section{Methodology}
\label{methodology}
\subsection{Problem Formulation}
Suppose the model receives a sequences of tasks $T=\{T_1,T_2,\dots,T_n\}$ over time, where each task $T_i$ consists of a set of input-output pairs $D_i = \{(x_{i1},y_{i1}), (x_{i2},y_{i2}), \dots, \\(x_{im},y_{im})\}$, where $y_i\cap y_{i'}=\emptyset$, for $i \neq i'$. For each task $T_i$, the model should produce a function $f_i:X\rightarrow Y$ that maps the input space $X$ to the output space $Y$. The goal of CIL is to learn a sequence $F=\{f_1, f_2,\dots, f_n\}$ such that the loss over all tasks is minimized. The expected risk of CIL is defined as:
\vspace{-0.5em}
\begin{equation}
    \centering
    f^* = \underset{f\in F}{\arg\min}\ \mathbb{E}_{(x,y)\sim T}
    \vspace{-0.4em}
\end{equation}

Crucially, when learning a new task $T_i$, the model should preserve its performance on the previous tasks $\{T_1,T_2,\dots, \\ T_{i-1}\}$, and catastrophic forgetting should be minimized or ideally, eliminated. Additionally, the model needs to handle a series of unlearning requests. Assuming the model has learned all classes in $C$ and the classes to be unlearned are $C_f$, then the model should preserve its discrimination ability for remaining classes $C_r=C \setminus C_f$.

\vspace{-1em}
\subsection{eCIL-MU Process}

Fig. \ref{eCIL-MU} illustrates the training phase and inference phase of eCIL-MU. For effectuating and validating our idea, we employ the pre-trained ResNet-50 as the embedding model $M_e$ parameterized by $w$. For a given input data $d$, $M_e$ performs a series of convolution and pooling operations to obtain vectors $v_l$ that represents the $d$'s features, in which $v_l=f_e(d; w)$, $f_e$ denotes the entire network architecture of $M_e$. 

During the CIL process, the acquired vector $V$ is stored in the vector database DB-CIL:$\{v_l\}$. For the MU process, in relation to the vectors $v$ corresponding to $C_f$, a matching is conducted with the vectors in DB-CIL. We employ cosine similarity to identify vectors within the same classes.



Upon computing the cosine similarity, we utilize the K-nearest neighbor (KNN) \cite{guo2003knn} to identify the class of unlearning data and select the most similar vectors $v$ with same class. Subsequently, $v$ is transferred from DB-CIL to DB-MU.
   
Meanwhile, we should consider the CIL process. Assume the model is parameterized by $\theta$, for newcoming data $x_i$ with label $y_i$, our objective function is to minimize the loss function $L$ for $T_i$:
\vspace{-0.5em}
\begin{equation}
    \centering
    \underset{\theta}{min}\ L(y_i, g(x_i, \theta))
    \vspace{-0.5em}
\end{equation}

where $g(\cdot)$ is the hypothesis function. As the performance of neural nets models relies on weights, the forward pass and back-propagation tends to prioritize updating weights for the current task's categories. This can result in a loss of grasp over previously learned knowledge and even lead to catastrophic forgetting. Therefore, concerning CIL, it is necessary to possess the capability both to acquire new knowledge and to preserve existing knowledge. In order to solve this stability-plasticity dilemma, we adopt the model SSRE \cite{zhu2022self} employing main branch knowledge distillation to transfer invariant knowledge, achieved by oversampling ($U_{P_B}$) prototypes with a batch size $B$:
\vspace{-0.5em}
\begin{equation}
    \centering
    P_B = U_{P_B}(Prototype)
    \vspace{-0.5em}
\end{equation}

A structural reorganization strategy is used as side branch that integrates structural expansion retaining old knowledge.

\vspace{-0.5em}
\begin{equation}
    \centering
    f_e^t(d; \theta^t) = f_e^{t-1}(d;\widehat{\theta}_e^{t-1} \oplus \Delta \theta^t)
    \vspace{-0.2em}
    \label{feature_embedding}
\end{equation}

where $\widehat{\theta}$ indicates fixed parameters and $\oplus$ represents the structural expansion operation. Here, zero-padding and linear transformation are used to integrate the parameters from the side branch with the model parameters.

As illustrated in Fig.\ref{eCIL-MU}, during the inference phase, the same embedding model $M_e$ is used as a filter before CIL-MU model to determine whether the input data $D_{in}$ belongs to $C_f$. To simplify, assuming the input vector $v_{in} = f_e(D_{in};w)$ and vectors in DB-MU $v_{f}$, giving a threshold $s$, we have:
\vspace{-0.5em}
\begin{equation}
    \centering
    D_{in} \in C_f \quad s.t.\quad cos(v_{in},v_{f}) \geq s
    \vspace{-0.5em}
    \label{threshold}
\end{equation}

For input data belonging to DB-CIL, we process it using CIL-MU and output its predictions. Otherwise, we have designed four output strategies and compare their effectiveness.

\vspace{0.5em}
\noindent{\textbf{Uniform random:}} Specifically, assuming the model has learned a total of $N$ classes, the output formula is as follows:
\vspace{-0.5em}
\begin{equation}
    \centering
    y =  Randint(N)
    \vspace{0.5em}
\end{equation}

\vspace{-1em}
\begin{figure}
    \centering
    \includegraphics[width=0.49\textwidth]{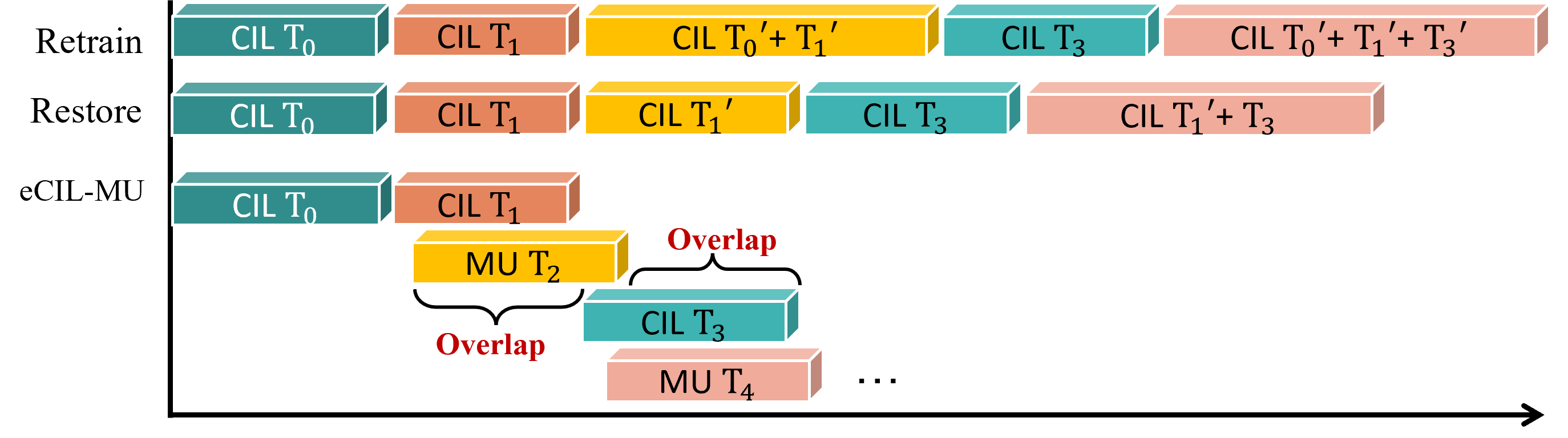}
    \vspace{-2em}
    \caption{Retraining from scratch and restoring and resuming training are serial processes. In contrast, eCIL-MU enables partial parallelization, allowing overlap once $M_e$ embeds $C_f$ or preceding CIL task ends.}
    \label{overlapping}
    \vspace{-1.5em}
\end{figure}

\noindent{\textbf{Proportional to all classes:}} Assuming the number of each class $i$ is $n_i$, then the probability of output being class $i$ is:

\vspace{-1em}
\begin{equation}
    \centering
    p(i) = \frac{n_i}{\sum_{j=1}^N n_j}
    \vspace{-0.5em}
\end{equation}

Sampling based on $p$ values prioritizes classes with larger counts, giving them higher selection probabilities than classes with smaller counts.

\vspace{0.5em}
\noindent{\textbf{Inversely proportional to distances:}} When the size of data in DB-CIL become substantial, calculating distances becomes complex, causing $\sum n_i$ data transfers from disk to memory. To simplify, we approximate distances by computing the cosine similarity between  $v_f$ and the centroids $c_i$ of each class. The following formula calculates probabilities:

\vspace{-0.5em}
\begin{equation}
    \centering
    p(i) = \frac{1}{cos(v_f, c_i)}
    \vspace{-0.5em}
\end{equation}

\vspace{0.5em}
\noindent{\textbf{Shift to the nearest class:}} Using Eq.\ref{threshold} for filtering introduces errors that result in certain vectors not being unlearned being incorrectly identified as unlearned vectors (seen Fig.\ref{cf}). These misidentified vectors are inhrently close to the $C_f$ vectors. We shift $C_f$ to the nearest class to rectify these misclassified classes and produce accurate outputs.  
\vspace{-0.5em}
\begin{equation}
    \centering
    y = y \sim max(cos(v_f, c_i))
    \vspace{-0.5em}
\end{equation}

As shown in Fig.\ref{overlapping}, when employing the eCIL-MU framework for MU tasks, once the embedding model completes the extraction of input data features, the next CIL or MU task can then be initiated. However, the subsequent CIL task must wait until the previous CIL task concludes.

\vspace{-0.5em}
\section{Experiment}
\label{sec:experiment}

\noindent{\textbf{Dataset:}} To demonstrate the performance of the proposed eCIL-MU framework, we conduct experiments using the CIFAR-10 and CIFAR-100 datasets \cite{krizhevsky2009learning}. CIFAR-10 consists of 5,000 training images and 1,000 testing images across 10 classes, with each image sized at $32\times32$ pixels. CIFAR-100, on the other hand, consists of 500 training images and 100 testing images for each of its 100 classes, also sized $32\times32$.

\vspace{0.5em}
\noindent{\textbf{Settings:}} We employ ResNet18 as our backbone model to re-implement SSRE \cite{zhu2022self}. Adam is used as optimizer with an initial learning rate 1e-3 and decay rate 5e-4. Training epoch is set to 100 and the batch size is 128. We utilize milvus to store the vectors we constructed. During the vector matching process, we set K=100 for KNN. The threshold $s$ in Eq.\ref{threshold} is set to 0.77 since we aim to achieve relatively high values for both true positive (TP) and true negative (TN) as well as low values of false positive (FP) and true negative (TN) at the corresponding $s$. All experiments are conducted in a single NVIDIA Quadro p6000 GPU. 

\vspace{0.5em}
\noindent{\textbf{Baseline:}} Retraining from scratch as well as restoring and resuming training are applied as baseline methods. For retraining from scratch, assuming that the data corresponding to the unlearning class exists within task $T_i$, we remove the data of unlearning class from $T_i$ and initiate the CIL process anew. For restoring and resuming training, we save intermediate model $M_i$ after each CIL. For MU, we restore from $M_i$, which lacks any unlearning class information, and conduct the CIL task using the remaining data.

\vspace{0.5em}
\noindent{\textbf{Metrics:}} In the eCIL-MU framework, both CIL and MU tasks are conducted. Hence, we need to measure the effects of CIL and MU. We consider the accuracy concerning the remaining classes $C_r$, and the accuracy related to unlearning classes $C_f$. Furthermore, as our method accommodates both CIL and MU simultaneously, we obtain training time and assess the acceleration compared to baseline methods, represented by the speedup ratio.

\vspace{-1em}
\begin{figure}[h!]
    \centering
    \includegraphics[width=0.45\textwidth]{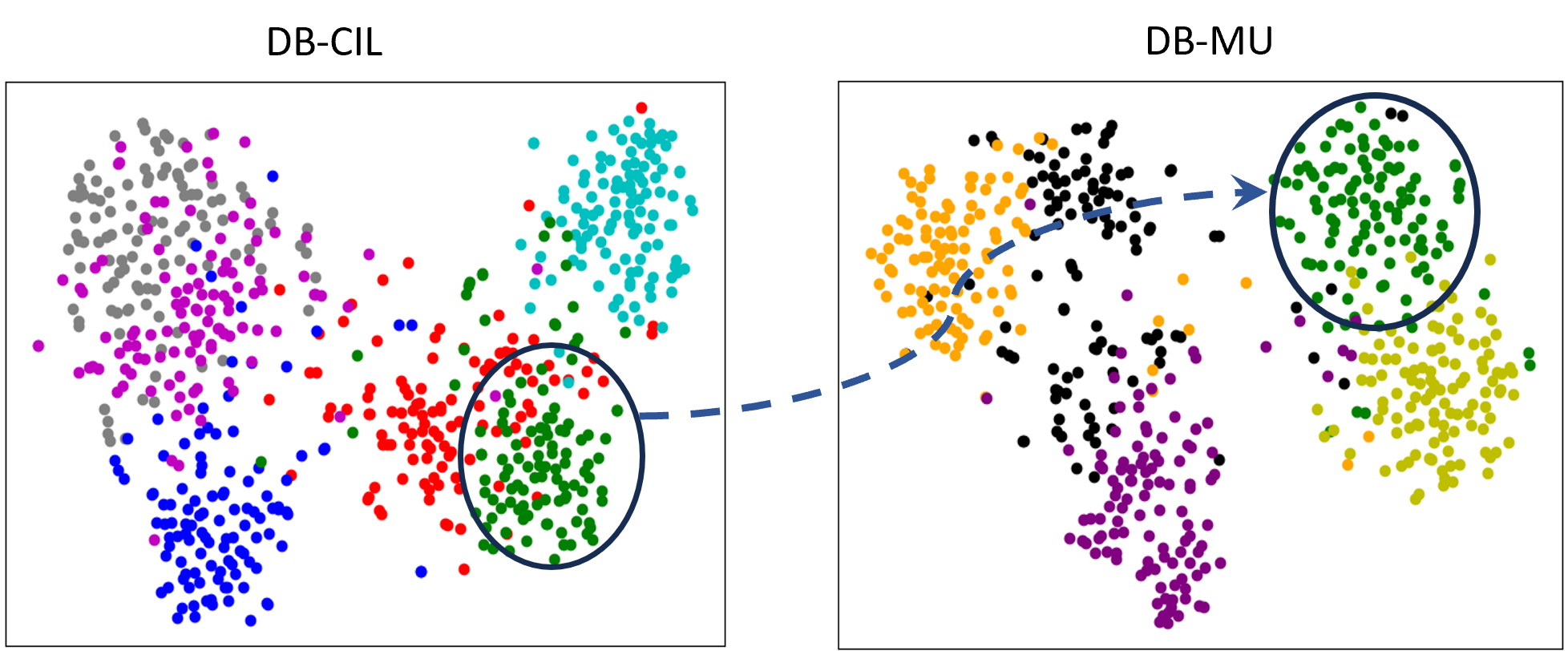}
    \vspace{-1em}
    \caption{t-SNEs of embedding vectors of CIFAR-10 mapped by $M_e$. Vectors in green belong to unlearning class, and will be transferred from DB-CIL to DB-MU.}
    \label{transfer}
    \vspace{-1em}
\end{figure}

\vspace{0.5em}
\noindent{\textbf{Experimental Results:}}  Fig.\ref{transfer} illustrates the t-SNE visualization after employing $M_e$ to embed the training data. The cluster in green represents the class $C_i$ to be unlearned. Utilizing vector matching, such cluster can be identified and transferred from the DB-CIL to the DB-MU. On the CIFAR-10, applying KNN achieves an accuracy of 87\%, which implies the vector migration is reasonably precise.

For CIFAR-10, we initialize the model with 5 classes to obtain model $M_0$($T_0$), achieving a test accuracy($Acc_{t}$) of 95.1\%. The following tasks are performed on $M_0$ sequentially: MU-1($T_1$), CIL-1($T_2$), MU-1($T_3$), CIL-1($T_4$), where MU-1 represents unlearning one class and CIL-1 signifies incremental learning one class. 

\begin{figure}[t]
\setlength{\abovecaptionskip}{-0.1cm}
\setlength{\belowcaptionskip}{-0.4cm}
\centering
\subfigure[]{
    \begin{minipage}[t]{0.3\linewidth}
    \label{cf}
    \centering
    \includegraphics[width=1\textwidth]{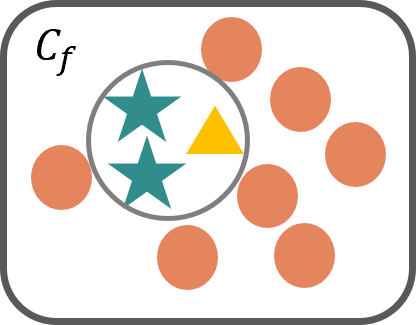}
    \end{minipage}
}
\subfigure[]{
    \begin{minipage}[t]{0.3\linewidth}
    \label{random}
    \centering
    \includegraphics[width=1\textwidth]{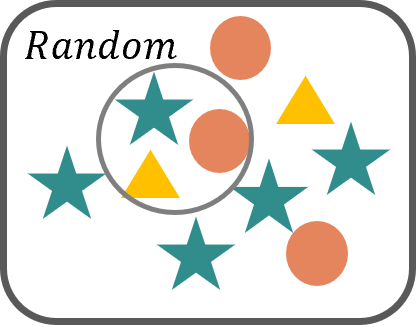}
    \end{minipage}
}%
\subfigure[]{
    \begin{minipage}[t]{0.3\linewidth}
    \label{nearest}
    \centering
    \includegraphics[width=1\textwidth]{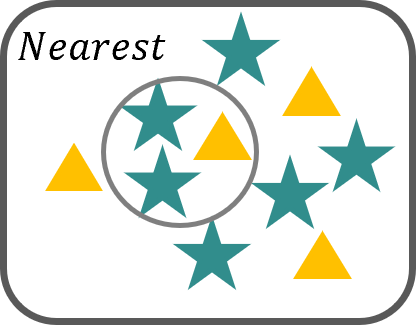}
    \end{minipage}
}%
\centering
\caption{(a) shows the predicted $C_f$ after filtering (via Eq.\ref{threshold}). Different symbols represent the ground-truth of these samples. Circles indicate classes in $C_f$, while others represent classes in $C_r$. (b) represents potential predictions for samples in (a) using various random methods. (c) illustrates predictions under the strategy shift-to-the-nearest-class.}
\vspace{-0.3em}
\end{figure}

Tab.\ref{results} shows the experimental results. The last method CIL indicates that the model solely undergoes CIL tasks without engaging in MU. For the first three strategies of our eCIL-MU, similar results are obtained under the same task. Ideally, the accuracy on $C_f$ for uniform random output should be around 20\% and the accuracy on $C_r$ should be around 80\%. From a theoretical perspective, after initializing $M_0$, performing a MU-1 task ($T_1$) to unlearn one class, with $s=0.77$, the testing results yield a recall (TP/(TP+FN)) of 0.9186 and a specificity (TN/(FP+TN)) of 0.709. That is to say, after $T_1$, the accuracy on $C_r$ should approximate $recall*Acc_t=87.35\%$. Moreover, due to the specificity of 0.709, 29.1\% the data in $C_f$ are misclassified as $C_r$. These misclassified data are predicted correctly with a confidence of 95.1\%, leading to an overall accuracy of approximately 41.85\% on $C_f$. The results in Tab.\ref{results} also corroborate this analysis.

\begin{table}[h!]
\vspace{-0.2em}
\footnotesize
    \begin{tabular}{@{\hspace{0.1em}}c@{\hspace{0.1em}}|c@{\hspace{0.3em}}c@{\hspace{0.9em}}c@{\hspace{0.3em}}c@{\hspace{0.9em}}c@{\hspace{0.3em}}c@{\hspace{0.9em}}c@{\hspace{0.3em}}c@{\hspace{0.9em}}c@{\hspace{0em}}}
    \hline
    \multicolumn{1}{c|}{\multirow{2}{*}{Method}}  & \multicolumn{2}{c}{MU-1} & \multicolumn{2}{c}{CIL-1} & \multicolumn{2}{c}{MU-1} & \multicolumn{2}{c}{CIL-1} & Time  \\ \cline{2-10} 
    \multicolumn{1}{l|}{}         & $C_r$            & $C_f$      & $C_r$            & $C_f$   & $C_r$            &$C_f$  &$C_r$            & $C_f$   \\ \hline
    Retrain  & 95.70  & 0   & 82.25     & 0   & 95.32    & 0      &  82.03    & 0  &5916          \\
    Restore   &  95.70   & 0   & 82.12   & 0   & 95.70    &  0    &  81.70     & 0  &3886        \\
    Random    & 83.73     & 41.30     &  70.10   &33.80  & 65.95    &  43.15  &  58.82  &  36.55  &1409     \\
    Proportional  & 83.43  & 41.80   & 69.98 & 35.50    & 65.93    & 43.55    & 58.76    &  36.20  &1409   \\
    Inverse    &    83.27  & 40.40  & 69.70    & 33.10 & 65.25   & 40.65  & 58.52   & 35.38    &1409        \\
    Nearest  & 94.45   & 27.00    & 79.82 & 21.70    & 79.57    &  32.90   & 72.04   & 27.05 &1409  \\ 
    CIL   &   -  & -    &    79.80   & -      &  -   &   -    &  70.09     & -    &1390        \\ \hline
    \end{tabular}
    \caption{Accuracy (\%) on $C_r$ and $C_f$ after CIL or MU tasks in which several classes are learned or unlearned and the total time taken ($s$). Higher accuracy on $C_r$ and lower accuracy on $C_f$ indicate a better joint CIL-MU performance.}
    \label{results}
    \vspace{-1em}
\end{table}

We see that the strategy shift-to-the-nearest-class outperforms others. The accuracy on $C_r$ is significantly higher compared to others (a highest difference of 14.32\% after $T_3$), approaching the results of CIL (no more than a 2\% difference) and the accuracy on $C_f$ is much lower (a highest difference of 14.8\% after $T_1$). As shown in Fig.\ref{random}, under different random strategies, some of the data belonging to $C_f$ could still be classified correctly (circles in the same position in Fig.\ref{cf} and Fig.\ref{random}). However in Fig.\ref{nearest}, for the misclassified data, shifting them to the nearest class results in their original labels (stars and triangles enclosed by gray circles in Fig.\ref{nearest} and \ref{cf}). On the other hand, the truly unlearned classes, lacking corresponding vectors in the DB-CIL, will inevitably be misclassified. Thus, leading to an increase in accuracy on $C_r$ and a decrease in accuracy on $C_f$. The comparable accuracy on $C_r$ to that achieved with CIL suggests that using a `shift-to-nearest-class' strategy during predictions helps offset errors made during the filtering process.

For both retraining and restoring and resuming training, their accuracy on $C_f$ is 0 , as the model has not learned from $C_f$. For the CIL model, with each CIL task, its accuracy tends to decrease. This is also manifested in retraining and restoring and resuming training, where their accuracy on $C_r$ exhibits periodic fluctuations in our setup.

As the model learns new classes, the declining accuracy in CIL implies an inevitable occurrence of catastrophic forgetting. However, using the first three random output strategies can further accelerate the process of catastrophic forgetting. Employing the shift-to-the-nearest-class strategy allows the model to maintain the same rate of catastrophic forgetting as CIL, thereby ensuring the model's robustness.

\vspace{-1em}
\begin{figure}[h]
\setlength{\abovecaptionskip}{-0.1cm}
\setlength{\belowcaptionskip}{-0.4cm}
\centering
\subfigure[]{
    \begin{minipage}[t]{0.47\linewidth}
    \label{time}
    \centering
    \includegraphics[width=1\textwidth]{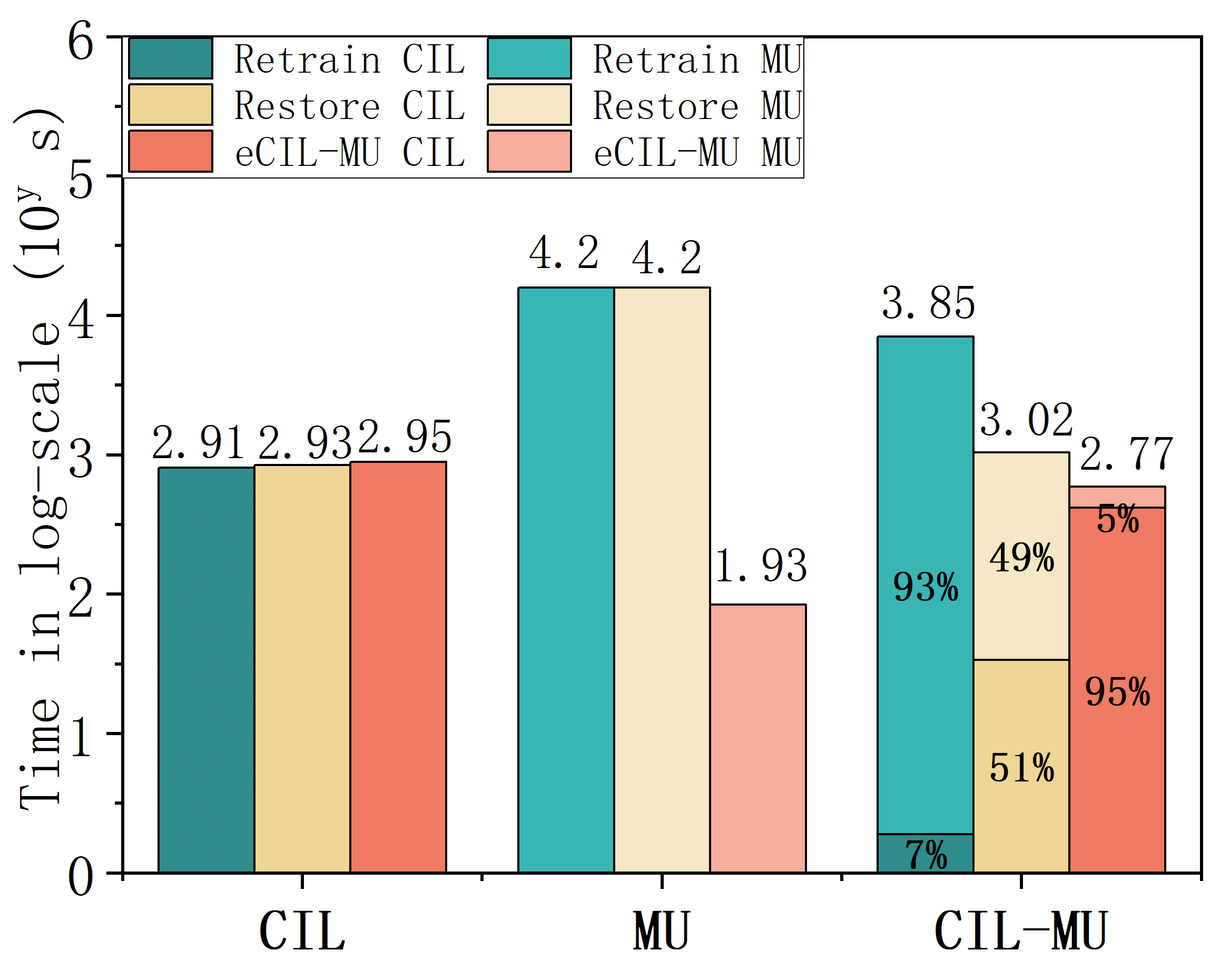}
    \end{minipage}
}
\subfigure[]{
    \begin{minipage}[t]{0.47\linewidth}
    \label{speedup}
    \centering
    \includegraphics[width=1\textwidth]{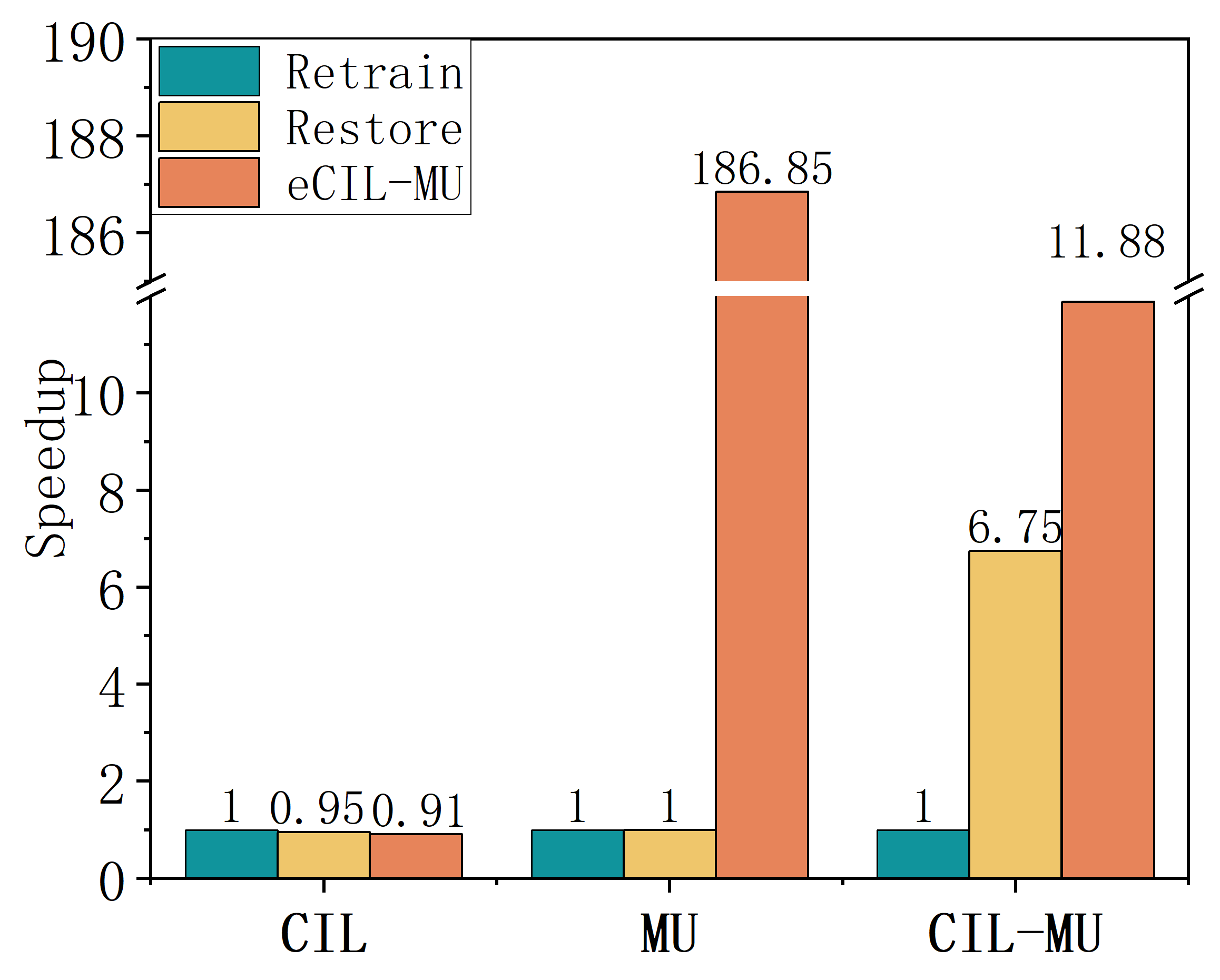}
    \end{minipage}
}%
\centering
\caption{(a): The time taken in log-scale ($10^y~s$) by each method in various task scenarios, as well as the proportion of CIL and MU task within the mixed CIL-MU. (b) illustrates the acceleration rates of restoring and resuming training, along with eCIL-MU, compared to retraining.  }
\vspace{0.5em}
\end{figure}

As shown in Fig.\ref{overlapping}, eCIL-MU allows partial overlap between CIL and MU tasks. The acceleration effect is illustrated by the time taken for each strategy in Tab.\ref{results}. For further illustration, we initialize CIL-50 using CIFAR-100 to obtain $M_0$ and test three following requests scenarios: 8 CIL tasks, 8 MU tasks, and a CIL-MU mix. Classes for unlearning in the mixed setup were randomly chosen. 

As shown in Fig.\ref{time}, in scenarios with exclusively CIL tasks, both restoring and resuming training, as well as eCIL-MU take longer than retraining due to model saving or embedding completion requirements. In cases of exclusively MU tasks, eCIL-MU is notably faster by avoiding constant retraining or restoration. As depicted in Fig.\ref{speedup}, it achieves an acceleration of up to 278.48$\times$. For mixed CIL and MU tasks, time distribution varies based on their relative proportions in the overall workload.

\vspace{-1em}
\section{Conclusion}
\label{sec:conclusion}
\vspace{-0.5em}
We combine Class Incremental Learning and Machine Unlearning to propose a non-destructive framework eCIL-MU, based on embedding techniques. Utilizing vector databases, it accomplishes CIL and MU tasks by modifying data. The overlap of CIL and MU accelerates model training. We identify that a shift-to-the-nearest-class strategy for unlearned classes enhances the model's robustness.

\label{sec:typestyle}

\bibliographystyle{IEEEbib}
\bibliography{main}

\end{document}